\documentclass[runningheads]{llncs}
\usepackage[T1]{fontenc}
\usepackage{graphicx}
\usepackage{amsmath}
\usepackage{multirow}
\usepackage[table,xcdraw]{xcolor}

\usepackage{microtype}
\usepackage{float}

\begin{document}
\title{nigam@COLIEE-22: Legal Case Retrieval and Entailment using Cascading of Lexical and Semantic-based models}

%
%
\author{Shubham Kumar Nigam$^{*}$\inst{1}, \and
Navansh Goel$^{*}$\inst{2}}
%
%
\institute{Indian Institute of Technology, Kanpur, India \\
\email{sknigam@iitk.ac.in}\\
\and
Vellore Institute of Technology, Chennai, India\\
\email{navansh.goel@gmail.com}}
%

\newcommand\SN[1]{\textbf{\textcolor{red}{(#1)$_{Shubham}$}}}

\newcommand\NavG[1]{\textbf{\textcolor{orange}{(#1)$_{Navansh}$}}}

\maketitle              
\def\thefootnote{*}\footnotetext{These authors contributed equally to this work}\def\thefootnote{\arabic{footnote}}

\begin{abstract}
This paper describes our submission to the Competition on Legal Information Extraction/Entailment 2022 (COLIEE-2022) workshop on case law competition for tasks 1 and 2. Task 1 is a legal case retrieval task, which involves reading a new case and extracting supporting cases from the provided case law corpus to support the decision. Task 2 is the legal case entailment task, which involves the identification of a paragraph from existing cases that entails the decision in a relevant case. We employed the neural models Sentence-BERT and Sent2Vec for semantic understanding and the traditional retrieval model BM25 for exact matching in both tasks. As a result, our team ("nigam") ranked 5th among all the teams in Tasks 1 and 2. Experimental results indicate that the traditional retrieval model BM25 still outperforms neural network-based models.

\end{abstract}
\section{Introduction}
Many countries, such as India, the United States, Canada, Australia, and Africa, follow the case law system. These case law systems are legal systems that give great weight to judicial precedent, and the volume of information produced in the legal sector nowadays is overwhelming. However, it takes significant efforts of legal practitioners to search for relevant cases from the database and extract the entailment parts manually with the rapid growth of digitalized legal documents. Therefore, an efficient legal assistant system to alleviate the heavy document work is necessary for Legal-AI.

The Competition on Legal Information Extraction and Entailment (COLIEE) workshop has been organized for several years to assist the legal research community. They provide the legal texts and specific problems in the legal domain such as question-answering, case law retrieval, case law entailment, statute law retrieval, and statute law entailment to help the research community develop Legal-AI techniques. 

This paper introduces our approaches to completing two case law tasks, Task 1 (the retrieval task) and Task 2 (the entailment task). We employed both traditional retrieval models like BM25\cite{robertson1995okapi} for exact matching and semantic understanding like Sentence-BERT\cite{reimers-gurevych-2019-sentence}, Sent2Vec\cite{pagliardini-etal-2018-unsupervised} in both task, and their combination in Task 1. In combination, we first select top-K candidates according to the BM25 rankings and afterward get the representation features of each sentence for the document we used pre-trained Sentence-BERT and Sent2Vec via comparing paragraph-level interaction. Furthermore, we used cosine similarity with the max-pooling strategy to get the final document score between query and noticed cases. As a result, our team ranked 5th among all the teams in both tasks. Experimental results suggest that the exact matching using the traditional retrieval model BM25 still outperforms representation features generated by neural network-based models. We released the codes for all subtasks via GitHub\footnote{\url{https://github.com/ShubhamKumarNigam/COLIEE-22}}.

\section{The Task}

\subsection{Task 1 - Legal Case Retrieval}
The main objective of legal case law retrieval is to obtain relevant supporting documents for a given query case $Q$. The task involves retrieving noticed cases $N = \{N1, N2, ..., N_m\} \mid N \subseteq S$ given that $S = \{S1, S2, S3, ..., S_n\}$ belonging to a given case law corpus of all possible supporting cases. The corpus predominantly consists of case law documents from the Federal Court of Canada database provided by Compass Law.

The query case documents contain suppressed reference markers instead of actual references to precedent case laws, compelling the participants to design models that can understand the text around the references and return valid noticed cases. A supporting case $S_k : S_k \in S$, is a noticed case for query $Q_j : Q_j \in Q$, \emph{iff} $Q_j$ contains a reference to $S_k$.

\subsection{Task 2 - Legal Case Entailment}
Task 2 comprises identifying a specific paragraph from a given supporting case that entails the decision for the query case. Thus, given a query case $Q$, a supporting case $R$ and paragraphs $P = \{P1, P2, P3, ...., Pn\}$ such that $P \subseteq R$, the task requires one to correctly identify $P_i \in P$ such that it entails the decision for query case $Q$.

\section{Data Corpus}
The data corpus for both Task 1 and Task 2 belongs to a database of case law documents from the Federal Court of Canada provided by Compass Law. Table \ref{tab:data-stats} presents the dataset statistics for both the tasks. For Task 1, 898 query cases were given against 3531 candidate cases as a part of the training data corpus. The test dataset had 300 query cases against 1263 candidate cases. For Task 2, 525 query cases were provided for training against 18,740 paragraphs. There were 100 query cases against 3278 candidate paragraphs as part of the testing dataset. The organizers provided the candidate paragraphs for a given query case separately. On average, there are 35.627 candidate paragraphs for each query case in the training dataset and 32.455 candidate paragraphs for each query case in the testing dataset. 

On further analysis, we found that the average number of relevant candidate cases for Task 1 was 4.684 cases for the training dataset and 4.21 relevant cases for the test dataset. As a part of our methodology, we predict the top 5 possible noticed cases for each query case based on these numbers. The average number of relevant paragraphs for Task 2 was a little over 1 paragraph, 1.14 paragraphs for training and 1.18 paragraphs for testing. These numbers show that most documents for Task 2 had their decisions entailed inside a single paragraph. The table also shows the average length of each document. The average number of words is higher in the testing query dataset than in the training query dataset for Task 1.

On the contrary, the average number of words per candidate document decreases from the training data to the testing data. The opposite trend is observed for Task 2, where the average number of words per query document decreases from training data to testing data. In contrast, the average number of words per candidate paragraph increases from training to testing. 

\begin{table}[]
\centering
\caption{Statistics of the training and test data for Task1 and Task2				}
\label{tab:data-stats}
\resizebox{\textwidth}{!}{%
\begin{tabular}{|l|cc|cc|}
\hline
                             & \multicolumn{2}{c|}{\textbf{Task 1}} & \multicolumn{2}{c|}{\textbf{Task 2}} \\ \hline
 & \multicolumn{1}{c|}{\textbf{Train}} & \textbf{Test} & \multicolumn{1}{c|}{\textbf{Train}} & \textbf{Test} \\ \hline
\# of queries                & \multicolumn{1}{c|}{898}      & 300      & \multicolumn{1}{c|}{525}      & 100      \\ \hline
\# of candidate cases/paragraphs         & \multicolumn{1}{c|}{3,531}      & 1,263      & \multicolumn{1}{c|}{18,740}      & 3,278      \\ \hline
avg \# of candidate paragraphs per query         & \multicolumn{1}{c|}{-}      & -      & \multicolumn{1}{c|}{35.627}      & 32.455      \\ \hline
avg \# of relevant candidates/paragraphs   & \multicolumn{1}{c|}{4.684}      & 4.210      & \multicolumn{1}{c|}{1.14}      & 1.18      \\ \hline
avg query length (words)     & \multicolumn{1}{c|}{28,866.73}      & 32,461.27      & \multicolumn{1}{c|}{24,966.55}      & 22,326.95      \\ \hline
avg candidate length (words) & \multicolumn{1}{c|}{30,113.09}      & 29,756.87      & \multicolumn{1}{c|}{636.08}      & 643.69      \\ \hline
\end{tabular}%
}
\end{table}

\subsection{Preprocessing}
Since the given documents are very long and represent the entire document singly is not efficient. Also, it is often noted that case judgments cite multiple cases and write multiple sentences about them. So we segment the document into meaningful sentences using the spacy library\footnote{https://spacy.io/usage/linguistic-features\#sbd} and retrieve relevant contents using the citation marker "FRAGMENT\_SUPPRESSED," which can say context around the citation. These sentences feed the models to make a corpus or generate distributed representations.

\subsection{Evaluation Metrics}
The evaluation metrics of Tasks 1 and 2 are precision, recall, and F-measure. All the metrics are micro-average, which means the evaluation measure is calculated using the results of all queries. Definition of these measures are as follows:

\begin{equation}
    Precision = \frac{\text{\# correctly retrieved cases(paragraphs) for all queries}}{\text{\# retrieved cases(paragraphs) for all queries}}
\end{equation}

\begin{equation}
    Recall = \frac{\text{\# correctly retrieved cases(paragraphs) for all queries}}{\text{\# relevant cases(paragraphs) for all queries}}
\end{equation}

\begin{equation}
    F-measure = \frac{\text{2 * Precision * Recall}}{\text{Precision + Recall}}
\end{equation}

\section{Our Methods}

\subsection{BM25}
In previous COILEE submissions, many participants Kim et al. \cite{kim2022legal}, Ma et al. \cite{ma2021retrieving}, Rosa et al. \cite{rosa2021yes}, Shao et al. \cite{shao2020thuir} and Shao et al. \cite{shao2020bert} in IJCAI-20 concluded that traditional bag-of-words IR models have competitive performances in legal case retrieval tasks. So our first preference is to try the exact matching-based BM25 model \cite{robertson1995okapi}, which is a probabilistic relevance model based on a bag-of-words approach. The score of a document $D$ given a query $Q$ which contains the words $q_1, ..., q_n$ is given by:

\begin{equation} 
\label{eq:bm25} \text{score}(D,Q) = \sum_{i=1}^{n} \text{IDF}(q_i) \cdot \frac{f(q_i, D) \cdot (k_1 + 1)}{f(q_i, D) + k_1 \cdot (1 - b + b \cdot \frac{|D|}{\text{avgdl}})} 
\end{equation}

where $f(q_i, D)$ is $q_i$'s term frequency in the document $D$, $|D|$ is the length of the document $D$ in words, and avgdl is the average document length in the text collection from which documents are drawn. $k_1$ and $b$ are free parameters.

\begin{equation} 
\text{IDF}(q_i) = ln(\frac{N-n(q_i)+0.5}{n(q_i)+0.5}+1)
\end{equation}

where $N$ is the total number of documents in the collection, and $n(q_i)$ is the number of documents containing $q_i$\footnote{https://en.wikipedia.org/wiki/Okapi\_BM25}. 

We tried BM25 using Rank-BM25 library\footnote{https://github.com/dorianbrown/rank\_bm25}, where the algorithms are taken from the paper Trotman et al. \cite{trotman2014improvements}. Although we are not getting good results, we tried to implement Okapi BM25 with sklearn's TfidfVectorizer\footnote{https://scikit-learn.org/stable/modules/generated/sklearn.feature\_extraction.text.TfidfVectorizer.html} which converts a collection of raw documents to a matrix of TF-IDF features. TfidfVectorizer has valuable features; some important feature descriptions are:
\begin{itemize}
    \item \textbf{stop\_words:} Either we can pass string `english' or a list containing stop words, all of which will be removed from the resulting tokens.
    \item \textbf{ngram\_range:} The lower and upper boundary of the range of n-values for different n-grams to be extracted. For example, an ngram\_range of (1, 1) means only unigrams, and (1, 2) means unigrams and bigrams.
    \item \textbf{max\_df:} While building the vocabulary, ignore terms with a document frequency strictly higher than the given threshold (corpus-specific stop words). The range is [0.0, 1.0].
    \item \textbf{min\_df:} While building the vocabulary, ignore terms with a document frequency strictly lower than the given threshold. This value is also called a cut-off in the literature. The range is [0.0, 1.0].
    \item \textbf{norm:} Each output row will have a unit norm, either:
    \begin{itemize}
        \item \textbf{`l2':} The sum of squares of vector elements is 1. The cosine similarity between two vectors is their dot product when the l2 norm has been applied.
        \item \textbf{`l1':} Sum of absolute values of vector elements is 1.
    \end{itemize}
\end{itemize}

These features are beneficial for BM25 to extract better features, and even we can restrain these features. Our results depict that after employing TfidfVectorizer, BM25 performed much better. 

\subsection{Sent2Vec}
Sent2Vec is an Unsupervised Learning of Sentence Embeddings or distributed representations of sentences using Compositional N-Gram Features. It can be thought of as an extension of word2vec (CBOW) to sentences. Sentence embedding is the average of the source word embeddings of its constituent words. We generate features from the pre-trained model "sent2vec\_wiki\_bigrams" from Sent2vec\footnote{https://github.com/epfml/sent2vec} library and paper Pagliardini et al. \cite{pagliardini-etal-2018-unsupervised}, which is trained on the English Wikipedia dataset and embedding dimensions are 700. Now to find the most similar sentences between query case and candidate case in a dataset, we compared using cosine-similarity\footnote{https://scikit-learn.org/stable/modules/generated/sklearn.metrics.pairwise.cosine\_similarity.html} using max-pool strategy. Finally, we used the max-pooling strategy between query and noticed cases to get the most similar document.

\subsection{Sentence-BERT}
Since we need sentence/paragraph representation, instead of using any transformer and generating sentence embedding using a pooling strategy, it will be more suitable to directly use a model to derive semantically meaningful representations for sentences. We use the Sentence-BERT\footnote{https://www.sbert.net/} framework based on the paper \cite{reimers-gurevych-2019-sentence}, which provides an easy method to compute dense vector representations for sentences and paragraphs. This model is based on transformer networks like BERT/RoBERTa, but this modification of the pre-trained BERT network uses siamese and triplet network structures to derive semantically meaningful sentence embeddings. We generate dense vectors from the pre-trained model "all-mpnet-base-v2" from Sentence-BERT\footnote{https://www.sbert.net/docs/pretrained\_models.html} library, which is trained on a large and diverse dataset of over 1 billion training pairs, max sequence length 384, mean-pooling, and generates 768 dimensions. To find the most similar paragraphs between query and candidate cases in a dataset, we compared using cosine-similarity using the max-pool strategy. Finally, we used the max-pooling strategy between query and noticed cases to get the most similar document.

\subsection{Reduced-Space}
\begin{figure}[h]
\includegraphics[width=\textwidth]{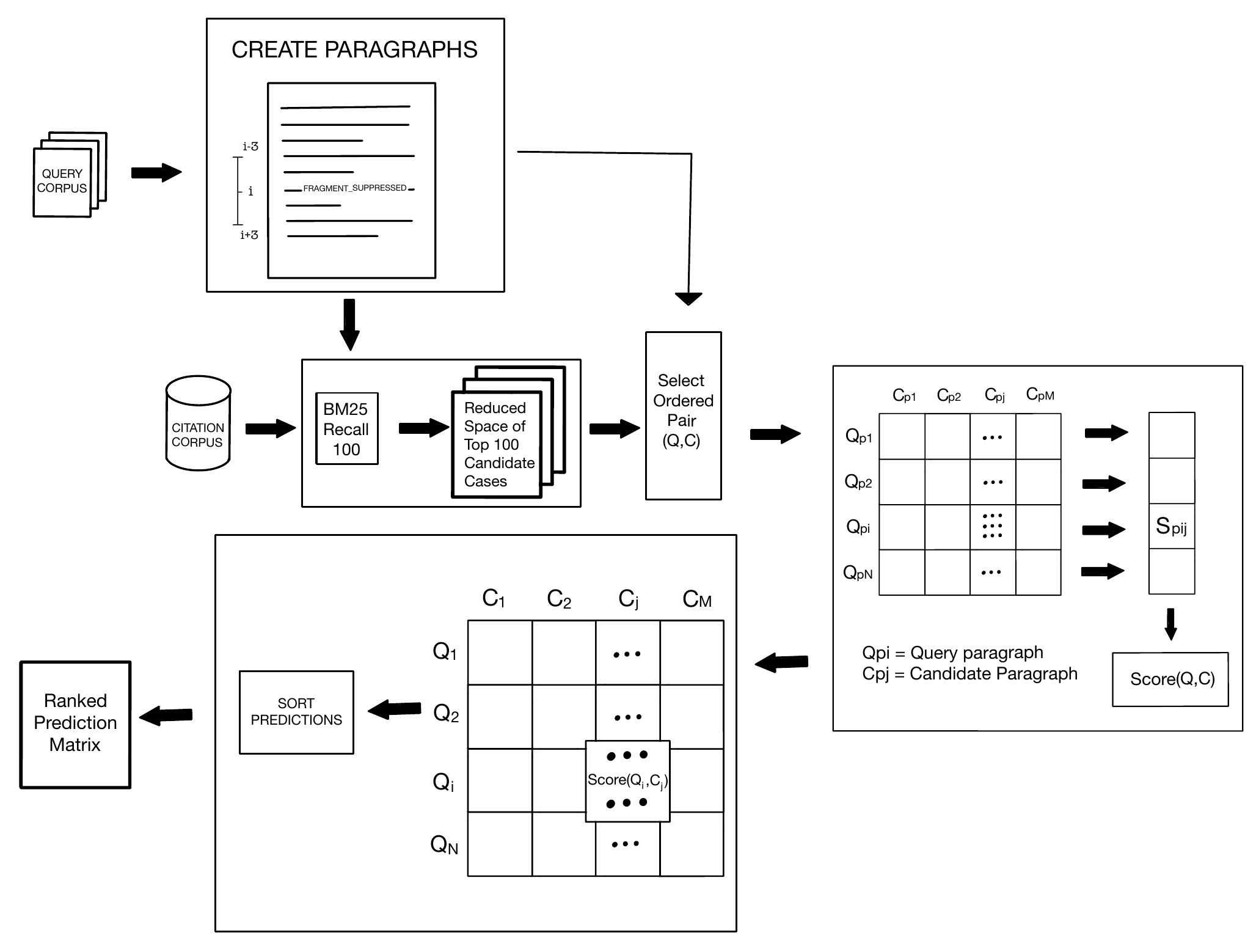}
\caption{An illustration of Reduced-Space Architecture.}
\label{fig:coliee-arch}
\end{figure}

As illustrated in Figure \ref{fig:coliee-arch}, we first represent the query corpus in paragraphs $Q = (Q_{p1}, Q_{p2}, ..., Q_{pN})$ by extracting the sentences which contain the "FRAGMENT\_SUPPRESSED" marker say $i$, three previous $i-3$ and three following sentences $i+3$ and citation corpus by segmentation of logical paragraphs $C = (C_{p1}, C_{p2}, ..., C_{pM})$. In total, we obtain $N*M$ pairs of paragraphs. Then, we used a lexical model to calculate the similarity between a given query and the entire case law corpus on paragraph-wise segmented documents. We limit the searching space from 3531 to 100 candidate cases by picking the top 100 cases with the highest BM25 scores for a given query. After reducing the searching space to 100 documents, we want to extract the semantic relationship between the query and candidate paragraphs pairs, e.g., $(Q_{pi}; C_{pj})$. To obtain this relationship score, we use the pre-trained model Sentence-BERT and Sent2vec. We used the max-pooling strategy to find the most similar paragraphs and documents in the same approach as above.

\section{Results and Analysis}
\subsection{Task-1}
We submitted three official runs in table \ref{tab:task-1-results}, run-1 is BM25, in a cascaded manner reduced space after using BM25 Recall@100, run-2 is pre-trained Sentence-BERT, and run-3 is pre-trained Sent2vec. All these official runs have done experiments on the paragraph level comparison, and every query case predicts a top-5 citation for both training and test dataset because the average number of relevant cases is approximately 5 in training data. We get the top score in using the traditional model BM25 for exact matching; on the other hand, in semantic understanding models, Sentence-BERT gets second, and Sent2vec gets third. Our team ("nigam") ranked 5th among all the teams in Tasks 1, and the results indicate that the traditional retrieval model BM25 still outperforms neural network-based models.

Apart from that, we tried several experiments without reduced space means the search space was the entire citation corpus in Sent2vec and Sentence-BERT. Similar behavior can be seen in table \ref{tab:task-1-training-results} for the training dataset, where BM25 performs better than Sentence-BERT and Sent2vec in reduced space. Along with experiments on the paragraph level comparison, we also tried document level comparison, but they did not give good results. We even included results without reduced space, and it can easily be observed that the reduced space results are slightly improved. Training and testing for every query case predict top-5 case law because the average number of relevant cases is approximately 5. We remove stop-words, and set parameter values max\_df=0.90, min\_df=1, b=0.99, k1=1.6, ngram\_range=(2,6). Moreover, other parameters are set to default for all experiments in BM25.

\begin{table}[h]
\centering
\caption{Results on the test set of Task 1}
\label{tab:task-1-results}
\resizebox{0.4\textwidth}{!}{%
\begin{tabular}{|c|c|c|c|}
\hline
\textbf{Team}  & \textbf{F1 Score} & \textbf{Precision} & \textbf{Recall} \\ \hline
UA             & 0.3715            & 0.4111             & 0.3389          \\
UA             & 0.371             & 0.4967             & 0.2961          \\
siat           & 0.3691            & 0.3005             & 0.4782          \\
siat           & 0.368             & 0.3026             & 0.4695          \\
UA             & 0.3559            & 0.363              & 0.3492          \\
siat           & 0.2964            & 0.2522             & 0.3595          \\
LeiBi          & 0.2923            & 0.3                & 0.285           \\
LeiBi          & 0.2917            & 0.2687             & 0.3191          \\
JNLP           & 0.2813            & 0.3211             & 0.2502          \\
\textbf{nigam} & \textbf{0.2809}   & \textbf{0.2587}    & \textbf{0.3072} \\
JNLP           & 0.2781            & 0.3144             & 0.2494          \\
DSSR           & 0.2657            & 0.2447             & 0.2906          \\
JNLP           & 0.2639            & 0.2446             & 0.2866          \\
DSSR           & 0.2461            & 0.2267             & 0.2692          \\
TUWBR          & 0.2367            & 0.1895             & 0.3151          \\
LeiBi          & 0.2306            & 0.2367             & 0.2249          \\
TUWBR          & 0.2206            & 0.1683             & 0.3199          \\
\textbf{nigam} & \textbf{0.1542}   & \textbf{0.142}     & \textbf{0.1686} \\
\textbf{nigam} & \textbf{0.1484}   & \textbf{0.1367}    & \textbf{0.1623} \\
DSSR           & 0.1317            & 0.1213             & 0.1441           \\ \hline
\end{tabular}%
}
\end{table}

\begin{table}[h]
\centering
\caption{Task 1 results in the training set. Every query case predicts top-5 case law because the average number of relevant cases is approximately 5. We remove stop-words, max\_df=0.90, min\_df=1, b=0.99, k1=1.6, ngram\_range=(2,6) in BM25. Moreover, other parameters are set to default for all experiments.}
\label{tab:task-1-training-results}
\resizebox{0.5\textwidth}{!}{%
\begin{tabular}{|c|c|c|c|}
\hline
\textbf{Model} & \textbf{F1 Score} & \textbf{Precision} & \textbf{Recall} \\ \hline
\textbf{BM25}  & \textbf{0.1957}   & \textbf{0.1895}    & \textbf{0.2023} \\ \hline
Sent2vec       & 0.0812            & 0.0786             & 0.0839          \\ \hline
SBERT          & 0.0844            & 0.0817             & 0.0873          \\ \hline
\begin{tabular}[c]{@{}c@{}}BM25 + Sent2vec \\ (Reduced Space)\end{tabular} & 0.1080 & 0.0961 & 0.1234 \\ \hline
\begin{tabular}[c]{@{}c@{}}BM25 + SBERT \\ (Reduced Space)\end{tabular}    & 0.1017 & 0.0984 & 0.1051 \\ \hline
\end{tabular}%
}
\end{table}

\subsection{Task-2}
We submitted two official runs in table \ref{tab:task-2-results}; both runs are based on the BM25 model. Run-1 uses only entailed fragment text as a query, and run-2 uses the filtered base case such that search entailed fragment text into the base case then input to the model as searched results along with previous and following sentences to capture the other relevant information. Every query case predicts one case law paragraph because the average number of relevant paragraphs is approximately 1.14 in the training dataset. Our team ("nigam") ranked 5th among all the teams in Tasks 2, and the results indicate that using only fragment text as the query will give better scores. Although, we got the highest Recall score among all participants. 

Similar behavior can be seen in table \ref{tab:task-2-training-results} for the training dataset, where BM25 performs better when only fragment text as the query and outputs only a single paragraph. Apart from that, we also tried two paragraphs prediction for every query; indeed, that gives a better Recall value, but on the other hand, precision values drop significantly. We also tried experiments where along with fragment text, we retrieved previous and following sentences from the base case after matching fragment text to capture the other relevant information. For example, $[-i,+j]$ means $i$ previous and $j$ following sentences from the matched fragment sentences. However, the results show that results drop as we include the other information. We kept stop-words, and set parameter values max\_df=0.65, min\_df=1, b=0.7, k1=1.6, ngram\_range=(1,1). Moreover, other parameters are set to default for all experiments in BM25. 


\begin{table}[h]
\centering
\caption{Results on the test set of Task 2}
\label{tab:task-2-results}
\resizebox{0.4\textwidth}{!}{%
\begin{tabular}{|c|c|c|c|}
\hline
\textbf{Team}  & \textbf{F1 Score} & \textbf{Precision} & \textbf{Recall} \\ \hline
NM             & 0.6783            & 0.6964             & 0.661           \\
NM             & 0.6757            & 0.7212             & 0.6356          \\
JNLP           & 0.6694            & 0.6532             & 0.6864          \\
JNLP           & 0.6612            & 0.6452             & 0.678           \\
jljy           & 0.6514            & 0.71               & 0.6017          \\
jljy           & 0.6514            & 0.71               & 0.6017          \\
JNLP           & 0.6452            & 0.6154             & 0.678           \\
jljy           & 0.633             & 0.69               & 0.5847          \\
NM             & 0.6325            & 0.6379             & 0.6271          \\
UA             & 0.5446            & 0.6105             & 0.4915          \\
UA             & 0.5363            & 0.7869             & 0.4068          \\
UA             & 0.4121            & 0.3049             & 0.6356          \\
\textbf{nigam} & \textbf{0.3204}   & \textbf{0.198}     & \textbf{0.839}  \\
\textbf{nigam} & \textbf{0.2104}   & \textbf{0.13}      & \textbf{0.5508} \\ \hline
\end{tabular}%
}
\end{table}


\begin{table}[H]
\centering
\caption{Task 2 results in the training set. Every query case predicts either 1 or 2 case law paragraphs because the average number of relevant paragraphs is approximately 1.14. We kept stop-words, max\_df=0.65, min\_df=1, b=0.7, k1=1.6, ngram\_range=(1,1). Moreover, other parameters are set to default for all experiments in BM25. Query cases are input such that either only entailed fragment text or search that text into the base case with previous and following sentences. }
\label{tab:task-2-training-results}
\resizebox{0.8\textwidth}{!}{%
\begin{tabular}{|
>{\columncolor[HTML]{FFFFFF}}c |
>{\columncolor[HTML]{FFFFFF}}c |
>{\columncolor[HTML]{FFFFFF}}c |
>{\columncolor[HTML]{FFFFFF}}c |
>{\columncolor[HTML]{FFFFFF}}c |}
\hline
\textbf{Query case Input} &
  \textbf{\begin{tabular}[c]{@{}c@{}}\# Prediction \\ per query\end{tabular}} &
  \textbf{F1 Score} &
  \textbf{Precision} &
  \textbf{Recall} \\ \hline
\cellcolor[HTML]{FFFFFF}                                              & \textbf{1} & \textbf{0.6228} & \textbf{0.6667} & \textbf{0.5843} \\ \cline{2-5} 
\multirow{-2}{*}{\cellcolor[HTML]{FFFFFF}\textbf{\begin{tabular}[c]{@{}c@{}}consider only \\ Entailed Fragment text\end{tabular}}} &
  2 &
  0.5106 &
  0.4010 &
  0.7028 \\ \hline
\cellcolor[HTML]{FFFFFF}                                              & 1          & 0.4359          & 0.4667          & 0.4090          \\ \cline{2-5} 
\multirow{-2}{*}{\cellcolor[HTML]{FFFFFF}using base case {[}-1,+1{]}} & 2          & 0.4172          & 0.3276          & 0.5743          \\ \hline
\cellcolor[HTML]{FFFFFF}                                              & 1          & 0.4110          & 0.4400          & 0.3856          \\ \cline{2-5} 
\multirow{-2}{*}{\cellcolor[HTML]{FFFFFF}using base case {[}-1,+2{]}} & 2          & 0.3857          & 0.3029          & 0.5309          \\ \hline
\cellcolor[HTML]{FFFFFF}                                              & 1          & 0.3790          & 0.4057          & 0.3556          \\ \cline{2-5} 
\multirow{-2}{*}{\cellcolor[HTML]{FFFFFF}using base case {[}-1,+3{]}} & 2          & 0.3663          & 0.2876          & 0.5042          \\ \hline
\cellcolor[HTML]{FFFFFF}                                              & 1          & 0.3594          & 0.3848          & 0.3372          \\ \cline{2-5} 
\multirow{-2}{*}{\cellcolor[HTML]{FFFFFF}using base case {[}-2,+3{]}} & 2          & 0.3469          & 0.2724          & 0.4775          \\ \hline
\cellcolor[HTML]{FFFFFF}                                              & 1          & 0.3488          & 0.3733          & 0.3272          \\ \cline{2-5} 
\multirow{-2}{*}{\cellcolor[HTML]{FFFFFF}using base case {[}-3,+3{]}} & 2          & 0.3457          & 0.2714          & 0.4758          \\ \hline
\end{tabular}%
}
\end{table}

\section{Conclusion}
We participated at COLIEE 2022 in Task 1 and Task 2, which allowed exploring information retrieval challenges in the legal case law retrieval and legal entailment. Our main objective was to combine traditional lexical retrieval models BM25 with dense passage retrieval models Sentence-BERT and Sent2vec for re-ranking the results. We show that the paragraph-level retrieval in the first stage outperforms the document-level retrieval. Also, overall ranking after combining lexical and semantic models improves the ranking of dense passage retrieval models alone for task-1. Although usually, neural network-based models perform best in the NLP tasks, especially transformers. Nevertheless, our results and previous COLIEE participants show that these pre-trained models do not work well in a specialized domain; traditional lexical retrieval models still perform better.

Furthermore, in task-2, using only fragment text as the query will give better scores, whereas accommodating other information from the base case only impairs the accuracy. Since we got the highest Recall score among all participants, we should also try the reduced space approach. In the future, we plan to investigate dense retrieval models domain-specific contextualized language models like LegalBERT \cite{chalkidis-etal-2020-legal} and graph neural networks \cite{yang2021legalgnn}.


%
%
\bibliographystyle{splncs04}
\bibliography{mybibliography}

\begin{thebibliography}{10}
\providecommand{\url}[1]{\texttt{#1}}
\providecommand{\urlprefix}{URL }
\providecommand{\doi}[1]{https://doi.org/#1}

\bibitem{chalkidis-etal-2020-legal}
Chalkidis, I., Fergadiotis, M., Malakasiotis, P., Aletras, N., Androutsopoulos,
  I.: {LEGAL}-{BERT}: The muppets straight out of law school. In: Findings of
  the Association for Computational Linguistics: EMNLP 2020. pp. 2898--2904.
  Association for Computational Linguistics, Online (Nov 2020).
  \doi{10.18653/v1/2020.findings-emnlp.261},
  \url{https://aclanthology.org/2020.findings-emnlp.261}

\bibitem{kim2022legal}
Kim, M.Y., Rabelo, J., Okeke, K., Goebel, R.: Legal information retrieval and
  entailment based on bm25, transformer and semantic thesaurus methods. The
  Review of Socionetwork Strategies pp. 1--18 (2022)

\bibitem{ma2021retrieving}
Ma, Y., Shao, Y., Liu, B., Liu, Y., Zhang, M., Ma, S.: Retrieving legal cases
  from a large-scale candidate corpus  (2021)

\bibitem{pagliardini-etal-2018-unsupervised}
Pagliardini, M., Gupta, P., Jaggi, M.: Unsupervised learning of sentence
  embeddings using compositional n-gram features. In: Proceedings of the 2018
  Conference of the North {A}merican Chapter of the Association for
  Computational Linguistics: Human Language Technologies, Volume 1 (Long
  Papers). pp. 528--540. Association for Computational Linguistics, New
  Orleans, Louisiana (Jun 2018). \doi{10.18653/v1/N18-1049},
  \url{https://aclanthology.org/N18-1049}

\bibitem{reimers-gurevych-2019-sentence}
Reimers, N., Gurevych, I.: Sentence-{BERT}: Sentence embeddings using {S}iamese
  {BERT}-networks. In: Proceedings of the 2019 Conference on Empirical Methods
  in Natural Language Processing and the 9th International Joint Conference on
  Natural Language Processing (EMNLP-IJCNLP). pp. 3982--3992. Association for
  Computational Linguistics, Hong Kong, China (Nov 2019).
  \doi{10.18653/v1/D19-1410}, \url{https://aclanthology.org/D19-1410}

\bibitem{robertson1995okapi}
Robertson, S.E., Walker, S., Jones, S., Hancock-Beaulieu, M.M., Gatford, M.,
  et~al.: Okapi at trec-3. Nist Special Publication Sp  \textbf{109}, ~109
  (1995)

\bibitem{rosa2021yes}
Rosa, G.M., Rodrigues, R.C., Lotufo, R., Nogueira, R.: Yes, bm25 is a strong
  baseline for legal case retrieval. arXiv preprint arXiv:2105.05686  (2021)

\bibitem{shao2020thuir}
Shao, Y., Liu, B., Mao, J., Liu, Y., Zhang, M., Ma, S.: Thuir@ coliee-2020:
  Leveraging semantic understanding and exact matching for legal case retrieval
  and entailment. arXiv preprint arXiv:2012.13102  (2020)

\bibitem{shao2020bert}
Shao, Y., Mao, J., Liu, Y., Ma, W., Satoh, K., Zhang, M., Ma, S.: Bert-pli:
  Modeling paragraph-level interactions for legal case retrieval. In: IJCAI.
  pp. 3501--3507 (2020)

\bibitem{trotman2014improvements}
Trotman, A., Puurula, A., Burgess, B.: Improvements to bm25 and language models
  examined. In: Proceedings of the 2014 Australasian Document Computing
  Symposium. pp. 58--65 (2014)

\bibitem{yang2021legalgnn}
Yang, J., Ma, W., Zhang, M., Zhou, X., Liu, Y., Ma, S.: Legalgnn: Legal
  information enhanced graph neural network for recommendation. ACM
  Transactions on Information Systems (TOIS)  \textbf{40}(2),  1--29 (2021)

\end{thebibliography}

\end{document}